\documentclass{article}

\usepackage[preprint,nonatbib]{neurips_2020}

\usepackage[utf8]{inputenc} 
\usepackage[T1]{fontenc}    
\usepackage{caption}
\usepackage{subcaption}
\usepackage{url}            
\usepackage{booktabs}       
\usepackage{amsfonts}       
\usepackage{nicefrac}       
\usepackage{microtype}      
\usepackage{graphicx}
\usepackage{appendix}
\usepackage{marvosym}

\usepackage{booktabs}
\usepackage{tikz}
\usepackage{amssymb}
\usepackage{pifont}
\usepackage{units}
\usepackage{booktabs,amsfonts,dcolumn} 
\usepackage{comment}
\usepackage{adjustbox}
\usepackage{amsmath,amssymb} 
\usepackage{color}
\usepackage[pagebackref=false,breaklinks,colorlinks=true]{hyperref}
\usepackage{array}
\usepackage{multicol}
\usepackage{wrapfig}
\usepackage{multirow}
\usepackage{colortbl}
\newcommand{\cmark}{\ding{51}}%
\newcommand{\xmark}{\ding{55}}%
\definecolor{gray9}{gray}{.9}
\definecolor{gray95}{gray}{.95}
\newcommand{\etal}{\textit{et al.}}
\definecolor{gray8}{gray}{.8}
\definecolor{gray85}{gray}{.85}
\def\ie{\emph{i.e.}}
\def\eg{\emph{e.g.}}

\def\etal{{\em et al.~}}

\newcommand\blfootnote[1]{%
\begingroup
\renewcommand\thefootnote{}\footnote{#1}%
\addtocounter{footnote}{-1}%
\endgroup
}
\title{BEVFormer: Learning Bird's-Eye-View Representation from Multi-Camera Images via Spatiotemporal Transformers}

\author{
Zhiqi Li$^{1,2*}$,
Wenhai Wang$^{2*}$,
Hongyang Li$^{2*}$,
Enze Xie$^3$,
\textbf{Chonghao Sima}$^{2}$, \\
\textbf{Tong Lu}$^{1}$,
\textbf{Yu Qiao}$^{2}$,
\textbf{Jifeng Dai}$^{2}$\textsuperscript{\Letter}
\\ [0.15cm]
$^1$Nanjing University~~~~$^2$Shanghai AI Laboratory~~~~
$^3$The University of Hong Kong
}

\begin{document}

\maketitle
\begin{figure}[h]
\centering
\includegraphics[width=0.9\linewidth]{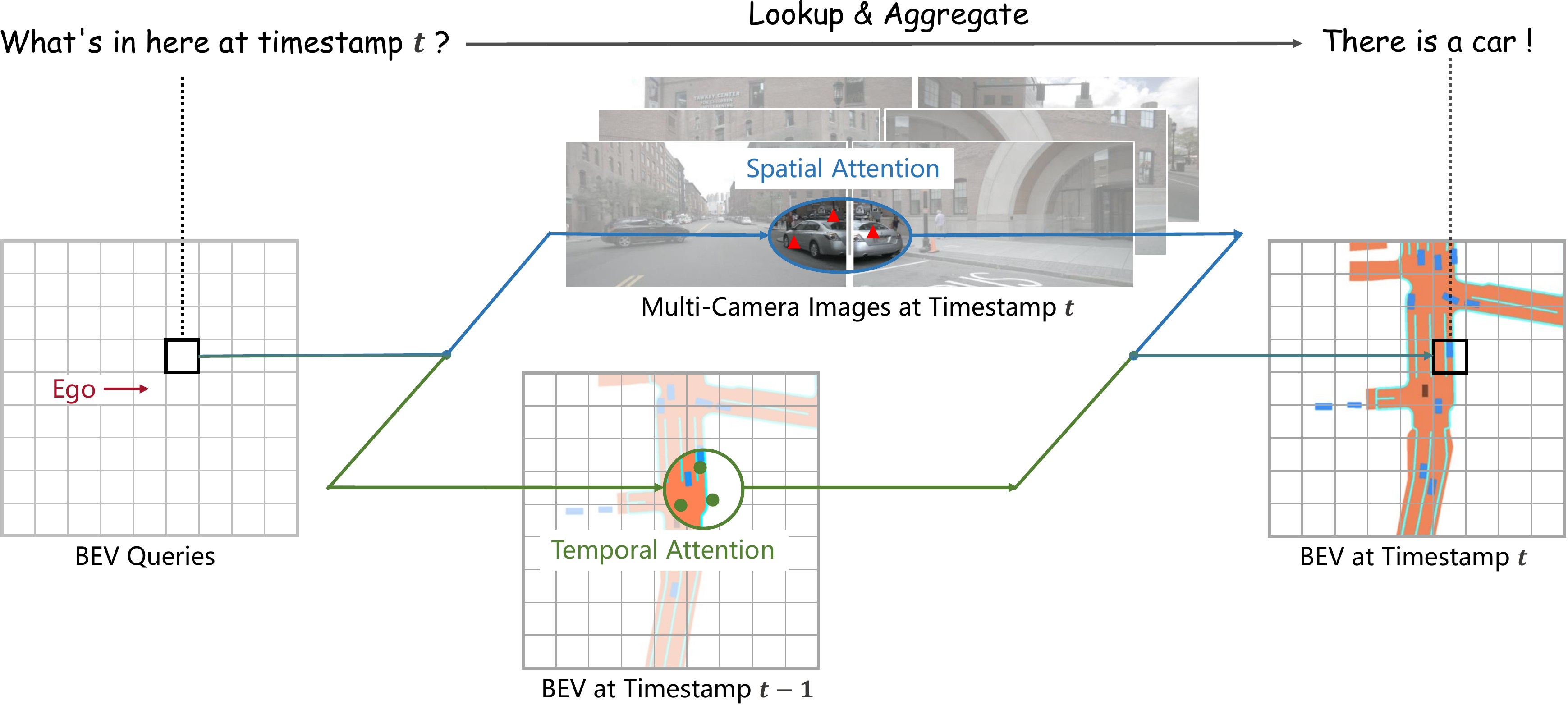}
\caption{We propose \textbf{BEVFormer}, a paradigm
for autonomous driving that applies 
both Transformer and Temporal structure to generate 
bird's-eye-view (BEV) features from multi-camera inputs. 
BEVFormer leverages 
queries to lookup spatial/temporal space and aggregate spatiotemporal information correspondingly, 
hence benefiting stronger representations for perception tasks.
}
\label{fig:moti}
\end{figure}

\begin{abstract}
\blfootnote{*: Equal contribution. This work is done when Zhiqi Li is an intern at Shanghai AI Lab.}
\blfootnote{\Letter: Corresponding author.}
3D visual perception tasks, including 3D detection and map segmentation based on multi-camera images, are essential for autonomous driving systems. In this work, we present a new framework termed BEVFormer, which learns unified BEV representations with spatiotemporal transformers to support multiple autonomous driving perception tasks. In a nutshell, BEVFormer exploits both spatial and temporal information by interacting with spatial and temporal space through predefined grid-shaped BEV queries. To aggregate spatial information, we design spatial cross-attention that each BEV query extracts the spatial features from the regions of interest across camera views. For temporal information, we propose temporal self-attention to recurrently fuse the history BEV information.
Our approach achieves the new state-of-the-art 56.9\% in terms of NDS metric on the nuScenes \texttt{test} set, which is 9.0 points higher than previous best arts and on par with the performance of LiDAR-based baselines. We further show that BEVFormer remarkably improves the accuracy of velocity estimation and recall of objects under low visibility conditions. The code is available at \url{https://github.com/zhiqi-li/BEVFormer}.

\end{abstract}

\section{Introduction}
Perception in 3D space is critical for various applications such as autonomous driving, robotics, \textit{etc}.
Despite the remarkable progress of LiDAR-based methods \cite{vora2020pointpainting,lang2019pointpillars,zhou2018voxelnet,yan2018second,chen2017multi},
camera-based approaches~\cite{wang2021fcos3d,philion2020lift,wang2022detr3d,pan2020cross} have attracted extensive attention in recent years.
Apart from the low cost for deployment, cameras own the desirable advantages to detect long-range distance objects and identify vision-based road elements (\textit{e.g.}, traffic lights, stoplines), compared to LiDAR-based counterparts.

Visual perception of the surrounding scene in autonomous driving is expected to predict the 3D bounding boxes or the semantic maps from 2D cues given by multiple cameras. The most straightforward solution is based on the  monocular frameworks \cite{wang2021fcos3d,wang2022probabilistic,park2021pseudo,reiher2020sim2real,bruls2019right} and cross-camera post-processing. The downside of this framework is that
it processes different views separately and cannot capture information across cameras, leading to 
low performance and efficiency~\cite{philion2020lift,wang2022detr3d}.

As an alternative to the monocular frameworks, a more unified framework is extracting holistic representations from multi-camera images. 
The bird's-eye-view (BEV) is a commonly used representation of the surrounding scene since it clearly presents the location and scale of objects and is suitable for various autonomous driving tasks, such as perception and planning~\cite{ng2020bev}.
Although previous map segmentation methods demonstrate BEV's effectiveness~\cite{philion2020lift,hu2021fiery,ng2020bev}, BEV-based approaches have not shown significant advantages over other paradigm in 3D object detections~\cite{wang2022detr3d,park2021pseudo,reading2021categorical}. 
The underlying reason is that the 3D object detection task requires strong BEV features to support accurate 3D bounding box prediction, but generating BEV from the 2D planes is ill-posed.
A popular BEV framework that generates BEV features is based on depth information~\cite{wang2019pseudo,philion2020lift,reading2021categorical}, but this paradigm is sensitive to the accuracy of depth values or the depth distributions. The detection performance of BEV-based methods is thus subject to compounding errors~\cite{wang2022detr3d}, and inaccurate BEV features can seriously hurt the final performance. Therefore, \emph{we are motivated to design a BEV generating method that does not rely on depth information and can learn BEV features adaptively rather than strictly rely on 3D prior.}
Transformer, which uses an attention mechanism to aggregate valuable features dynamically, meets our demands conceptually.

Another motivation for using BEV features to perform perception tasks is that BEV is a desirable bridge to connect temporal and spatial space. 
For the human visual perception system, temporal information plays a crucial role in inferring the motion state of objects and identifying occluded objects, and many works in vision fields have demonstrated the effectiveness of using video data~\cite{brazil2020kinematic,ma20223d,luo2018fast,qi2021offboard,kang2016object}.
However, the existing state-of-the-art multi-camera 3D detection methods rarely exploit temporal information.
The significant challenges are that autonomous driving is time-critical and objects in the scene change rapidly, and thus simply stacking BEV features of cross timestamps brings extra computational cost and interference information, which might not be ideal.
Inspired by recurrent neural networks (RNNs)~\cite{hochreiter1997long,cho2014properties}, \emph{we utilize the BEV features to deliver temporal information from past to present recurrently, which has the same spirit as the hidden states of RNN models.}

To this end, we present a transformer-based bird's-eye-view (BEV) encoder, termed \textbf{BEVFormer}, which can effectively aggregate spatiotemporal features from multi-view cameras and history BEV features.
The BEV features generated from the BEVFormer can simultaneously support multiple 3D perception tasks such as 3D object detection and map segmentation, which is valuable for the autonomous driving system.
As shown in Fig. \ref{fig:moti}, our BEVFormer contains three key designs, which are
(1) grid-shaped BEV queries to  fuse spatial and temporal features via attention mechanisms flexibly,
(2) spatial cross-attention module to aggregate the spatial features from multi-camera images,
and (3) temporal self-attention module to extract temporal information from history BEV features, which benefits the velocity estimation of moving objects and the detection of heavily occluded objects, while bringing negligible computational overhead. 
With the unified features generated by BEVFormer,
the model can collaborate with different task-specific heads such as Deformable DETR \cite{zhu2020deformable} and mask decoder \cite{li2021panoptic}, for end-to-end 3D object detection and map segmentation.

Our main contributions are as follows:

$\bullet$ We propose BEVFormer, a spatiotemporal transformer encoder that  projects multi-camera and/or timestamp input to BEV representations. With the unified BEV features, our model can simultaneously support multiple autonomous driving perception tasks, including 3D detection and map segmentation.
 
$\bullet$ We designed learnable BEV queries along with a spatial cross-attention layer and a temporal self-attention layer to lookup spatial features from cross cameras and temporal features from history BEV, respectively, and then aggregate them into unified BEV features.

$\bullet$ We evaluate the proposed BEVFormer on multiple challenging benchmarks, including nuScenes~\cite{caesar2020nuscenes} and Waymo~\cite{sun2020scalability}.
Our BEVFormer consistently achieves improved performance compared to the prior arts.
For example, under a comparable parameters and computation overhead, BEVFormer achieves 56.9\% NDS on nuScenes \texttt{test} set, outperforming previous best detection method DETR3D~\cite{wang2022detr3d} by 9.0 points (56.9\% \emph{vs.} 47.9\%). For the map segmentation task, we also achieve  the state-of-the-art performance, more than 5.0 points higher than Lift-Splat~\cite{philion2020lift} on the most challenging lane segmentation.
We hope this straightforward and strong framework can serve as a new baseline for following 3D perception tasks.

\section{Related Work}

\subsection{Transformer-based 2D perception}
Recently, a new trend is to use transformer to reformulate detection and segmentation tasks~\cite{carion2020end,zhu2020deformable,li2021panoptic}. 

DETR~\cite{carion2020end} uses a set of object queries to generate detection results by the cross-attention decoder directly.
However, the main drawback of DETR is the long training time. Deformable DETR~\cite{zhu2020deformable}  solves this problem by proposing deformable attention.
Different from vanilla global attention in DETR, the deformable attention interacts with local regions of interest, which only samples $K$ points near each reference point and calculates attention results, resulting in high efficiency and significantly shortening the training time. The deformable attention mechanism is calculated by:
\begin{equation}
\text{DeformAttn}(q, p, x) = \sum_{i=1}^{N_\text{head}} \mathcal{W}_i\sum_{j=1}^{N_\text{key}} \mathcal{A}_{ij} \cdot \mathcal{W}'_i x(p+ \Delta p_{ij}),
\label{eq:single_deform_attn_fun}
\end{equation}
where $q$, $p$, $x$ represent the query, reference point and input features, respectively. $i$ indexes the attention head, and $N_\text{head}$ denotes the total number of attention heads.
$j$ indexes the sampled keys, and $N_\text{key}$ is the total sampled key
number for each head.
$W_i\!\in\!\mathbb{R}^{C\times (C / H_{\rm head})}$ and $W'_i\!\in\!\mathbb{R}^{(C / H_{\rm head})\times C}$ are the learnable weights, where $C$ is the feature dimension. $A_{ij}\!\in\![0,1]$ is the predicted attention weight, and is normalized by $\sum_{j=1}^{N_\text{key}} A_{ij}\!= \!1$. $\Delta p_{ij}\!\in\!\mathbb{R}^2$ are the predicted offsets to the reference point $p$. $x(p + \Delta p_{ij})$ represents the feature at location $p + \Delta p_{ij}$, which is 
extracted by bilinear interpolation as in Dai~\etal\cite{dai2017deformable}.
In this work, we extend the deformable attention to 3D perception tasks, to efficiently aggregate both spatial and temporal information.

\subsection{Camera-based 3D Perception}
Previous 3D perception methods typically perform 3D object detection or map segmentation tasks independently.
For the 3D object detection task, early methods are similar to 2D detection methods ~\cite{brazil2019m3d,mousavian20173d,xu2018multi,Simonelli_2019_ICCV,zhou2019objects}, which usually predict the 3D bounding boxes based on 2D bounding boxes.
Wang~\etal\cite{wang2021fcos3d} follows an advanced 2D detector FCOS~\cite{tian2019fcos} and directly predicts 3D bounding boxes for each object. DETR3D~\cite{wang2022detr3d} projects learnable 3D queries in 2D images, and then samples the corresponding features for end-to-end 3D bounding box prediction without NMS post-processing. 
Another solution is to transform image features into BEV features and predict 3D bounding boxes from the top-down view.  Methods transform image features into BEV features with the depth information from  depth estimation~\cite{wang2019pseudo} or  categorical depth distribution~\cite{reading2021categorical}.
OFT~\cite{Roddick2019OrthographicFT} and ImVoxelNet~\cite{rukhovich2022imvoxelnet} project the predefined voxels onto image features to generate the voxel representation of the scene. Recently, M$^2$BEV~\cite{xie2022m} futher explored the feasibility of simultaneously performing multiple perception tasks based on BEV features.

Actually, generating BEV features from multi-camera features is more extensively studied in map segmentation tasks~\cite{philion2020lift,pan2020cross}. A straightforward method is converting perspective view into the BEV through Inverse Perspective Mapping (IPM)~\cite{reiher2020sim2real,can2021structured}. In addition, Lift-Splat~\cite{philion2020lift} generates the BEV features based on the depth distribution.
Methods~\cite{pan2020cross,hendy2020fishing,chitta2021neat} utilize multilayer perceptron to learn the translation from perspective view to the BEV.
PYVA~\cite{yang2021projecting} proposes a cross-view transformer that converts the front-view monocular image into the BEV, but this paradigm is not suitable for fusing multi-camera features due to the computational cost of global attention mechinism~\cite{vaswani2017attention}. In addition to the spatial information, previous works~\cite{hu2021fiery,saha2021translating,can2020understanding} also consider the temporal information  by stacking BEV features from several timestamps. Stacking BEV features constraints the available temporal information within fixed time duration and brings extra computational cost.
In this work, the proposed spatiotemporal transformer generates BEV features of the current time by considering both spatial and temporal clues, and the temporal information is obtained from the previous BEV features by the RNN manner, which only brings little computational cost.

\section{BEVFormer}
Converting multi-camera image features to bird's-eye-view (BEV) features can provide a unified surrounding environment representation for various autonomous driving perception tasks.
In this work, we present  a new transformer-based framework for BEV generation, which can effectively aggregate spatiotemporal features from multi-view cameras and history BEV features via attention mechanisms.

\begin{figure}[t]
\centering
\includegraphics[width=\linewidth]{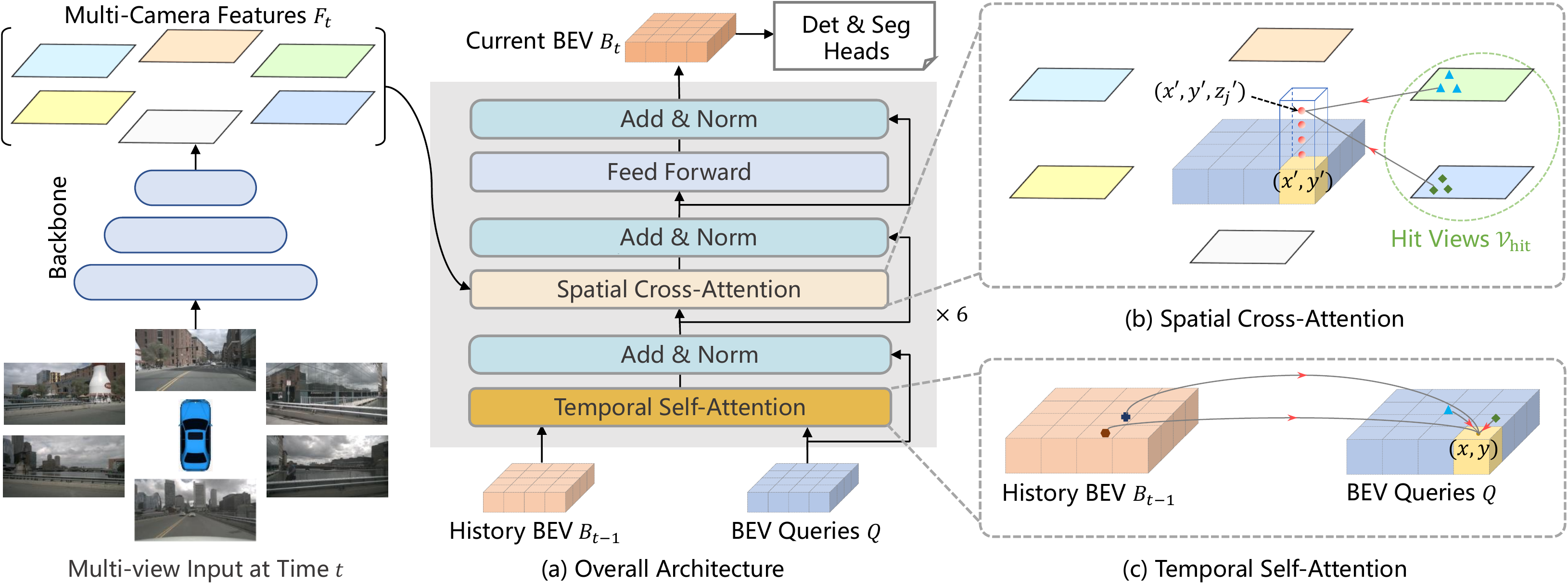}
\caption{\textbf{Overall architecture of BEVFormer.} (a) The encoder layer of BEVFormer contains grid-shaped BEV queries, temporal self-attention, and spatial cross-attention. (b) In spatial cross-attention, each BEV query only interacts with image features in the regions of interest. (c) In temporal self-attention, each BEV query interacts with two features: the BEV queries at the current timestamp and the BEV features at the previous  timestamp.}
\label{fig:model}
\end{figure}

\subsection{Overall Architecture}
As illustrated in Fig.~\ref{fig:model}, BEVFormer has 6 encoder layers, each of which follows the conventional structure of transformers~\cite{vaswani2017attention}, except for three tailored designs, namely BEV queries, spatial cross-attention, and temporal self-attention.
Specifically, BEV queries are grid-shaped learnable parameters, which is designed to query features in BEV space from multi-camera views via attention mechanisms. Spatial cross-attention and temporal self-attention are attention layers working with BEV queries, which are used
to lookup and aggregate spatial features from multi-camera images as well as temporal features from history BEV, according to the BEV query.

During inference, at timestamp $t$, we feed multi-camera 
images to the backbone network (\eg, 
ResNet-101~\cite{he2016deep}), and obtain the features 
$F_t\!=\!\{F_t^i\}_{i=1}^{N_{\rm view}}$ of different camera 
views, where $F_t^i$ is the feature of the $i$-th view, 
$N_{\rm view}$ is the total number of camera views. 
At the same time, we preserved the BEV features $B_{t\!-\!1}$ at the prior timestamp $ t\!-\!1$.
In each encoder layer, we first use BEV queries $Q$ to query the temporal information from the prior BEV features $B_{t\!-\!1}$ via the temporal self-attention.
We then employ BEV queries $Q$ to inquire about the spatial information from the multi-camera features $F_t$ via the spatial cross-attention. After the feed-forward network~\cite{vaswani2017attention}, the encoder layer output the refined BEV features, which is the input of the next encoder layer.
After 6 stacking encoder layers, unified BEV features $B_t$ at current timestamp $t$ are generated. Taking the BEV features $B_t$ as input, the 3D detection head and map segmentation head predict the  perception results such as 3D bounding boxes and semantic map.

\subsection{BEV Queries}\label{bev_queries}
We predefine a group of grid-shaped learnable parameters $Q\!\in\! \mathbb{R}^{H\!\times\!W\!\times\!C}$ as the queries of BEVFormer, where $H,W$ are the spatial shape of the BEV plane. 
To be specific, the query $Q_p\!\in\!\mathbb{R}^{1\!\times\!C} $ located at $p=(x,y)$ of $Q$ is responsible for the corresponding grid cell region in the BEV plane. 
Each grid cell in the BEV plane corresponds to a real-world size of $s$ meters. 
The center of BEV features corresponds to the position of the ego car by default.
Following common practices~\cite{gehring2017convolutional}, we add learnable positional embedding
to BEV queries $Q$ before inputting  them to BEVFormer.

\subsection{Spatial Cross-Attention}\label{spatial_attention}
Due to the large input scale of multi-camera 3D perception (containing $N_\text{view}$ camera views), the computational cost of vanilla multi-head attention~\cite{vaswani2017attention} is extremely high.
Therefore, we develop the spatial cross-attention based on deformable attention~\cite{zhu2020deformable}, which is a resource-efficient attention layer where each BEV query $Q_p$ only interacts with its regions of interest  across camera views. 
However, deformable attention is originally designed for 2D perception, so some adjustments are required for 3D scenes.

As shown in Fig.~\ref{fig:model} (b),
we first lift each query on the BEV plane to
a pillar-like query~\cite{lang2019pointpillars}, sample ${N_\text{ref}}$ 3D reference points from the pillar, and then project these points to 2D views. For one BEV query, the projected 2D points can only fall on some views, and other views are not hit. Here, we term the hit views as $\mathcal{V}_\text{hit}$.
After that, we regard these 2D points
as the reference points of the query $Q_p$ and sample the features from the hit views $\mathcal{V}_\text{hit}$ around these
reference points.
Finally, we perform a weighted sum of the sampled features as the output of spatial cross-attention.
The process of spatial cross-attention (SCA) can be formulated as:
\begin{align}\label{sca}
    \text{SCA}(Q_p, F_t) &= \frac{1}{|\mathcal{V}_\text{hit}|} \sum_{i\in \mathcal{V}_\text{hit}} \sum_{j=1}^{{N_\text{ref}}}
    \text{DeformAttn}(Q_p, \mathcal{P}(p,i,j), F_t^i),
\end{align}
where $i$ indexes the camera view, $j$ indexes the reference points, and ${N_\text{ref}}$ is the total reference points for each BEV query. $F_t^i$ is the features of the $i$-th camera view. For each BEV query $Q_p$, we use a project function $\mathcal{P}(p, i, j)$ to get the $j$-th reference point on the $i$-th view image.

Next, we introduce how to obtain the reference points on the view image from the projection  function $\mathcal{P}$.
We first calculate the real world location $(x', y')$ corresponding to the query $Q_p$ located at $p=(x,y)$ of $Q$ as Eqn. \ref{eqn:real}.
\begin{equation}{\label{eqn:real}}
    x'\!=\! (x\!-\!\frac{W}{2})\!\times\! s; \quad y'\! = \!(y\!-\!\frac{H}{2})\!\times\! s,
\end{equation}

where $H$, $W$ are the spatial shape of BEV queries, $s$ is the size of resolution  of BEV's grids, and $(x',y')$ are the coordinates where the position of ego car is the origin. In 3D space, the objects located at $(x', y')$ will appear at the height of $z'$ on the z-axis. So we predefine a set of anchor heights $\{z'_j\}_{j=1}^{N_\text{ref}}$ to make sure we can capture clues that appeared at different heights.  
In this way, for each query $Q_p$, we obtain a pillar of 3D reference points ${(x', y', z'_j)}_{j=1}^{{N_\text{ref}}}$. Finally, we project the 3D reference points to different image views through the projection  matrix of cameras, which can be written as:
\begin{equation}
\begin{split}
\label{project}
\mathcal{P}(p,i,j)& = (x_{ij},y_{ij})\\
\text{where}\ z_{ij}\cdot\begin{bmatrix} x_{ij} &y_{ij} & 1 \end{bmatrix}^T&= T_i \cdot \begin{bmatrix} x' &y' &z'_j & 1 \end{bmatrix}^T .
\end{split}
\end{equation}
Here, $\mathcal{P}(p,i,j)$ is the 2D point on $i$-th view projected from $j$-th 3D point $(x', y', z'_j)$, $T_i\!\in\! \mathbb{R}^{3\!\times\! 4}$ is the known projection  matrix of the $i$-th camera.

\subsection{Temporal Self-Attention}\label{temporal_attention}
In addition to spatial information, temporal information is also crucial for the visual system to understand the surrounding environment~\cite{ma20223d}. 
For example, it is challenging to infer the velocity of moving objects or detect highly occluded objects from static images without temporal clues.
To address this problem, we design temporal self-attention, which can represent the current environment by incorporating history BEV features.

Given the BEV queries $Q$ at current timestamp $t$ and history BEV features $B_{t\!-\!1}$ preserved at timestamp $t\!-\!1$, we first align $B_{t\!-\!1}$ to $Q$ according to ego-motion to make the features at the same grid correspond to the same real-world location. 
Here, we denote the aligned history BEV features $B_{t\!-\!1}$ as $B'_{t\!-\!1}$. However, from times $t-1$ to $t$, movable objects travel in the real world with various offsets. It is challenging to construct the precise association of the same objects between the BEV features of different times. 
Therefore, we model this temporal connection between features through the temporal self-attention (TSA) layer, which can be written as follows:
\begin{align}\label{MSDeformAttn}
    \text{TSA}(Q_p, \{Q, B'_{t-1}\}) = 
    \sum_{V\in \{Q, B'_{t-1}\}} \text{DeformAttn}(Q_p, p, V),
\end{align}
where $Q_p$ denotes the BEV query located at $p=(x, y)$. In addition, different from the vanilla deformable attention, the offsets $\Delta p$ in temporal self-attention are predicted by the concatenation of $Q$ and $B'_{t-1}$.
Specially, for the first sample of each sequence, the temporal self-attention will degenerate into a self-attention without temporal information, where we replace the BEV features $\{Q, B'_{t-1}\}$ with duplicate BEV queries $\{Q, Q\}$.

Compared to simply stacking BEV in~\cite{hu2021fiery,saha2021translating,can2020understanding}, our temporal self-attention can more effectively model long temporal dependency. BEVFormer extracts temporal information from the previous BEV features rather than multiple stacking BEV features, thus requiring less computational cost and suffering less disturbing information.

\subsection{Applications of BEV Features}
Since the BEV features $B_t\in \mathbb{R}^{H\times W \times C}$  is a versatile 2D feature map that can be used for various autonomous driving perception tasks, the 3D object detection and map segmentation task heads can be developed based on 2D perception methods~\cite{zhu2020deformable,li2021panoptic} with minor modifications. 

\noindent\textbf{For 3D object detection}, we design an end-to-end 3D detection head based on the 2D detector Deformable DETR~\cite{zhu2020deformable}. The modifications include using single-scale BEV features $B_t$ as the input of the decoder, predicting 3D bounding boxes and velocity rather than 2D bounding boxes, and only using $L_1$ loss
to supervise 3D bounding box regression. With the detection head, our model can end-to-end predict 3D bounding boxes and velocity without the NMS post-processing.

\noindent\textbf{For map segmentation}, we design a map segmentation head based on a 2D segmentation method Panoptic SegFormer~\cite{li2021panoptic}. Since the map segmentation based on the BEV is basically the same as the common semantic segmentation, we utilize the mask decoder of ~\cite{li2021panoptic} and class-fixed queries to target each semantic category, including the car, vehicles, road (drivable area), and lane.

\subsection{Implementation Details}
\noindent\textbf{Training Phase.} 
For each sample at timestamp $t$, we randomly sample another 3 samples from the consecutive sequence of the past 2 seconds, and this random sampling strategy can augment  the diversity of ego-motion~\cite{zhu2017deep}. We denote the timestamps of these four samples as $t\!-\!3$, $t\!-\!2$, $t\!-\!1$ and $t$.
For the samples of the first three timestamps, they are responsible for recurrently generating the BEV features $\{B_{t-3}, B_{t-2}, B_{t-1}\}$ and this phase requires no gradients. For the first sample at timestamp $t\!-\!3$, there is no previous BEV features, and temporal self-attention degenerate into self-attention. 
At the time $t$, the model generates the BEV features $B_t$ based on both multi-camera inputs and the prior BEV features $B_{t-1}$, so that $B_t$ contains the temporal and spatial clues crossing the four samples. Finally, we feed the BEV features $B_t$ into the detection and segmentation heads and compute the corresponding loss functions.

\noindent\textbf{Inference Phase.} During the inference phase, 
we evaluate each frame of the video sequence in chronological order.
The BEV features of the previous timestamp are saved and used for the next, and this online inference strategy is time-efficient and consistent with practical applications. Although we utilize temporal information, our inference speed is still comparable with other methods~\cite{wang2021fcos3d,wang2022detr3d}.

\section{Experiments}
\subsection{Datasets}
We conduct experiments on two challenging public autonomous driving datasets, namely nuScenes dataset~\cite{caesar2020nuscenes} and Waymo open dataset~\cite{sun2020scalability}.

\noindent\textbf{The nuScenes dataset}~\cite{caesar2020nuscenes} contains 1000 scenes of roughly 20s duration each, and the key samples are annotated at 2Hz. Each sample consists of RGB images from 6 cameras and has 360° horizontal FOV. For the detection task, there are  1.4M annotated 3D bounding boxes from 10 categories. We follow the settings in~\cite{philion2020lift}  to perform BEV segmentation task. 
This dataset also provides the official evaluation metrics for the detection task. The mean average precision (mAP) of nuScenes is computed using the center distance on the ground plane rather than the 3D Intersection over Union (IoU) to match the predicted results and ground truth. The nuScenes metrics also contain 5 types of true positive metrics (TP metrics), including ATE, ASE, AOE, AVE, and AAE for measuring translation, scale, orientation, velocity, and attribute errors, respectively. 
The nuScenes also defines a nuScenes detection score (NDS)
as ${\rm NDS}\!=\!\frac{1}{10}[5{\rm mAP}\!+\!\sum_{{\rm mTP}\in \mathbb{TP}} (1\!-\!{\rm min}(1,{\rm mTP}))]$ to capture all aspects of the nuScenes detection tasks.

\noindent\textbf{Waymo Open Dataset}~\cite{sun2020scalability}  is a large-scale autonomous driving dataset with 798 training sequences and 202 validation sequences. Note that the five images at each frame provided by Waymo have only about 252° horizontal FOV, but the provided annotated labels are 360° around the ego car. We remove these bounding boxes that can not be visible on any images in training and validation sets. Due to the Waymo Open Dataset being large-scale and high-rate~\cite{reading2021categorical}, we use a subset of the training split by sampling every 5$^{th}$ frame from the training sequences and only detect the vehicle category. 
We use the thresholds of 0.5 and 0.7 for 3D IoU to compute the mAP on Waymo dataset. 

\subsection{Experimental Settings}\label{details}
Following previous methods~\cite{wang2021fcos3d,wang2022detr3d,park2021pseudo}, we adopt two types of backbone: ResNet101-DCN~\cite{he2016deep,dai2017deformable} that initialized from FCOS3D~\cite{wang2021fcos3d} checkpoint, and VoVnet-99~\cite{lee2019energy} that initialized from DD3D~\cite{park2021pseudo} checkpoint. By default, we utilize the output multi-scale features from FPN~\cite{Lin2017FeaturePN} with sizes of \nicefrac{1}{16}, \nicefrac{1}{32}, \nicefrac{1}{64} and the dimension of $C\!=\!256$ . For experiments on nuScenes, the default size of BEV queries is $200\!\times\!200$, the perception ranges are [$-$51.2m, 51.2m] for the $X$ and $Y$ axis and the size of resolution $s$ of BEV’s grid is 0.512m. We adopt learnable positional embedding for BEV queries.
The BEV encoder contains 6 encoder layers and constantly refines the BEV queries in each layer. The input BEV features $B_{t\!-\!1}$ for each encoder layer are the same and require no gradients. For each local query, during the spatial cross-attention module implemented by deformable attention mechanism, it corresponds to ${N_\text{ref}}\!=\!4$ target points with different heights in 3D space, and the predefined height anchors are sampled uniformly from $-5$ meters to 3 meters. For each reference point on 2D view features, we use four sampling points around this reference point for each head.
By default, we train our models with 24 epochs, a learning rate of $2
\!\times\! 10^{-4}$.

For experiments on Waymo, we change a few settings. Due to the camera system of Waymo can not capture the whole scene around the ego car~\cite{sun2020scalability}, the default spatial shape of BEV queries is $300\!\times\!220$, the perception ranges are [$-$35.0m, 75.0m] for the $X$-axis and  [$-$75.0m, 75.0m] for the $Y$-axis. The size of resolution $s$ of each gird is 0.5m. The ego car is at (70, 150) of the BEV. 

\noindent\textbf{Baselines.} To eliminate the effect of task heads and compare other BEV generating methods fairly, we use VPN~\cite{pan2020cross} and Lift-Splat~\cite{philion2020lift} to replace our BEVFormer and keep task heads and other settings the same.  We  also adapt BEVFormer into a static model called \textbf{BEVFormer-S} via adjusting the temporal self-attention into a vanilla self-attention without using history BEV features.

\begin{table}[t]
\begin{center}
\caption{
\textbf{3D detection results on nuScenes \texttt{test} set.} $*$ notes that VoVNet-99 (V2-99)~\cite{lee2019energy} was pre-trained on the depth estimation task with extra data~\cite{park2021pseudo}. ``BEVFormer-S'' does not leverage temporal information in the BEV encoder. ``L'' and ``C'' indicate LiDAR and Camera, respectively.
}

\setlength{\tabcolsep}{0.28mm}
\begin{tabular}{l  c c |c c| c c c c c c }
\toprule
Method &  Modality & Backbone & NDS$\uparrow$  & mAP$\uparrow$  & mATE$\downarrow$     & mASE$\downarrow$     & mAOE$\downarrow$     & mAVE$\downarrow$    & mAAE$\downarrow$    \\
\midrule

SSN~\cite{zhu2020ssn} &L &-& 0.569 &0.463 &-&-&-&-&-\\
CenterPoint-Voxel~\cite{yin2021center} & L &-& 0.655 & 0.580  & - &- &- &- &- \\
PointPainting~\cite{vora2020pointpainting}& L$\!\&\!$C&-&0.581 &0.464 &0.388&0.271&0.496&0.247&0.111\\
\midrule

FCOS3D~\cite{wang2021fcos3d}  & C  & R101& 0.428 &0.358& 0.690& 0.249& 0.452& 1.434 &\textbf{0.124} \\
PGD~\cite{wang2022probabilistic}  & C &  R101&0.448 &0.386& \textbf{0.626}& \textbf{0.245} &0.451 &1.509 &0.127 \\
\rowcolor{gray95}
BEVFormer-S    & C & R101& 0.462& 0.409 & 0.650 & 0.261 & 0.439 & 0.925 & 0.147 \\
\rowcolor{gray9}
BEVFormer    & C & R101& \textbf{0.535}& \textbf{0.445} & 0.631 & 0.257 & \textbf{0.405} & \textbf{0.435} & 0.143 \\
\midrule
DD3D~\cite{park2021pseudo}  & C &V2-99$^*$& 0.477 &0.418& \textbf{0.572} & \textbf{0.249} &\textbf{0.368} &1.014 &\textbf{0.124} \\

DETR3D~\cite{wang2022detr3d}  & C &  V2-99$^*$& 0.479 & 0.412 &0.641& 0.255 &0.394 &0.845& 0.133\\
\rowcolor{gray95}
BEVFormer-S& C &V2-99$^*$ & 0.495& 0.435 & 0.589&0.254 & 0.402 &0.842 & 0.131 \\
\rowcolor{gray9}
BEVFormer& C &V2-99$^*$ & \textbf{0.569}& \textbf{0.481} & 0.582&0.256 & 0.375 & \textbf{0.378} & 0.126 \\

\bottomrule
\end{tabular}
\label{main_det}
\end{center}
\end{table}
\begin{table}[t]
\begin{center}
\caption{
\textbf{3D detection results on nuScenes \texttt{val} set.} ``C'' indicates Camera.
}

\setlength{\tabcolsep}{0.87mm}
\begin{tabular}{l  c c |c c| c c c c c c  }
\toprule
Method &  Modality & Backbone & NDS$\uparrow$  & mAP$\uparrow$  & mATE$\downarrow$     & mASE$\downarrow$     & mAOE$\downarrow$     & mAVE$\downarrow$    & mAAE$\downarrow$    \\
\midrule

FCOS3D~\cite{wang2021fcos3d}&  C & R101& 0.415 & 0.343 & 0.725 & 0.263 & 0.422 & 1.292  &\textbf{0.153}  \\
PGD~\cite{wang2022probabilistic} & C &R101 & 0.428 &0.369 &0.683 &\textbf{0.260}& 0.439 &1.268 &0.185 \\
DETR3D~\cite{wang2022detr3d}& C& R101 & 0.425 &0.346 &0.773 &0.268 &0.383 &0.842 &0.216\\


\rowcolor{gray95}
BEVFormer-S & C & R101& 0.448 & 0.375 & 0.725 & 0.272 & 0.391 & 0.802 & 0.200\\
\rowcolor{gray9}
BEVFormer   & C & R101& \textbf{0.517}& \textbf{0.416} & \textbf{0.673} & 0.274 & \textbf{0.372} & \textbf{0.394} & 0.198 \\
\bottomrule
\end{tabular}
\label{val_det}
\end{center}
\end{table}

\begin{table}[t] 
\begin{center}
\caption{\textbf{3D detection results  on Waymo \texttt{val} set under  Waymo evaluation metric and nuScenes evaluation metric.} ``L1'' and ``L2'' refer ``LEVEL\_1'' and ``LEVEL\_2'' difficulties of Waymo~\cite{sun2020scalability}. *:  Only use the
front camera and only consider object labels in the front camera’s field of view (50.4°).
${\text{\textdagger}}$: We compute the NDS score by setting ATE and AAE to be 1. ``L'' and ``C'' indicate LiDAR and Camera, respectively.
}

\setlength{\tabcolsep}{0.5mm}
\begin{tabular}{l| c| c c| c c|c c c c c }
\toprule
\multirow{3}{*}{Method} &\multirow{3}{*}{Modality}  & \multicolumn{4}{c|}{Waymo Metrics} & \multicolumn{5}{c}{Nuscenes Metrics} \\
\cline{3-11}
&  &\multicolumn{2}{c|}{IoU=0.5} & \multicolumn{2}{c|}{IoU=0.7} & 
\multirow{2}{*}{NDS$^{\text{\textdagger}}$$\uparrow$} &
\multirow{2}{*}{AP$\uparrow$} & \multirow{2}{*}{ATE$\downarrow$} & 
\multirow{2}{*}{ASE$\downarrow$} & \multirow{2}{*}{AOE$\downarrow$}\\
&  & L1/APH & L2/APH  & L1/APH & L2/APH  \\
 \midrule
 PointPillars~\cite{lang2019pointpillars}& L & 0.866 & 0.801 & 0.638 & 0.557&0.685 & 0.838 & 0.143 & 0.132 & 0.070\\
 \midrule
 DETR3D~\cite{wang2022detr3d}&C & 0.220 & 0.216 & 0.055 & 0.051 & 0.394&0.388 & 0.741 & \textbf{0.156} & 0.108\\
 \rowcolor{gray9}
 BEVFormer&C & \textbf{0.280} & \textbf{0.241} & \textbf{0.061} & \textbf{0.052}& \textbf{0.426}
 & \textbf{0.440} & \textbf{0.679} & 0.157 & \textbf{0.101}\\
\midrule
CaDNN$^*$~\cite{reading2021categorical}&C & 0.175 & 0.165 & 0.050 & 0.045&-& - &- &- &-\\ 
\rowcolor{gray9}
BEVFormer$^*$&C & 0.308 & 0.277 & 0.077 & 0.069 &-&-&-&-&-\\
\bottomrule

\end{tabular}
\label{table:Waymo}
\end{center}
\end{table}

\begin{table}[t] 
\begin{center}

\caption{
\textbf{3D detection and map segmentation results on nuScenes \texttt{val} set.}
Comparison of training segmentation and detection tasks jointly or not.
*: We use VPN~\cite{pan2020cross} and Lift-Splat~\cite{philion2020lift} to replace our BEV encoder for comparison, and the task heads are the same. ${\text{\textdagger}}$: Results from their paper.
}

\setlength{\tabcolsep}{2.9mm}
\begin{tabular}{l| c c | c c | c c c c }
\toprule
\multirow{2}{*}{Method}& \multicolumn{2}{c|}{Task Head}
 & \multicolumn{2}{c|}{3D Detection} & \multicolumn{4}{c}{BEV Segmentation (IoU)} \\
&Det& Seg& NDS$\uparrow$ &mAP$\uparrow$  & Car & Vehicles & Road & Lane\\
\midrule
Lift-Splat$^{\text{\textdagger}}$~\cite{philion2020lift} &\xmark &\cmark &- &-& 32.1 & 32.1 & 72.9 &20.0\\
FIERY$^{\text{\textdagger}}$~\cite{hu2021fiery}&\xmark &  \cmark& - &-&- & 38.2 & - & -\\
\midrule
VPN$^*$~\cite{pan2020cross} & \cmark & \xmark & 0.333 & 0.253   &- & - &- &-\\
VPN$^*$ &\xmark &\cmark & - & -  & 31.0 & 31.8 &76.9 &19.4 \\
VPN$^*$ &\cmark & \cmark& 0.334 & 0.257  &36.6 & 37.3 & 76.0 & 18.0\\
Lift-Splat$^*$& \cmark &\xmark & 0.397 & 0.348  & -&-&-&-\\
Lift-Splat$^*$ &\xmark &\cmark &- &- &42.1 & 41.7 &77.7 & 20.0 \\
Lift-Splat$^*$& \cmark & \cmark&  0.410 & 0.344&  43.0 & 42.8 & 73.9 & 18.3\\
\rowcolor{gray95}
BEVFormer-S&\cmark & \xmark & 0.448 & 0.375&  -& - & - & -\\
\rowcolor{gray95}
BEVFormer-S& \xmark& \cmark  &- &-&43.1  &43.2  &\textbf{80.7} &21.3 \\
\rowcolor{gray95}
BEVFormer-S&\cmark & \cmark & 0.453 & 0.380 &  44.3& 44.4 & 77.6 & 19.8\\

\rowcolor{gray9}
BEVFormer&  \cmark  &\xmark & 0.517 & \textbf{0.416} &- & - & - & -\\
\rowcolor{gray9}
BEVFormer&\xmark & \cmark& - &-&44.8 & 44.8 & 80.1 & \textbf{25.7}\\
\rowcolor{gray9}
BEVFormer&\cmark & \cmark & \textbf{0.520} & 0.412 & \textbf{46.8} & \textbf{46.7} & 77.5 & 23.9\\

\bottomrule

\end{tabular}
\label{table:multi-tasks}
\end{center}
\end{table}

\subsection{3D Object Detection Results}

We train our model on the detection task with the detection head only for fairly comparing with previous state-of-the-art 3D object detection methods.
In Tab.~\ref{main_det} and Tab.~\ref{val_det}, we report our main results on nuScenes  \texttt{test} and \texttt{val} splits.  Our method outperforms previous best method DETR3D~\cite{wang2022detr3d} over 9.2 points  on \texttt{val} set (51.7\% NDS \emph{vs.~}42.5\% NDS), under fair training strategy and comparable model scales. On the \texttt{test} set, our model achieves 56.9\% NDS without bells and whistles, 9.0 points higher than DETR3D (47.9\% NDS). Our method can even achieve comparable performance 
to some LiDAR-based baselines such as SSN (56.9\% NDS)~\cite{zhu2020ssn} and PointPainting (58.1\% NDS)~\cite{vora2020pointpainting}.

Previous camera-based methods~\cite{wang2022detr3d,park2021pseudo,wang2021fcos3d} were almost unable to estimate the velocity, and our method demonstrates that temporal information plays a crucial role in velocity estimation for multi-camera detection.  
The mean Average Velocity Error (mAVE) of BEVFormer is 0.378 m$/$s on the \texttt{test} set, outperforming other camera-based methods by a vast margin and approaching the performance of LiDAR-based methods~\cite{vora2020pointpainting}.

We also conduct experiments on Waymo, as shown in Tab.~\ref{table:Waymo}. Following~\cite{reading2021categorical}, we evaluate the  vehicle category with IoU criterias of 0.7 and 0.5. In addition, We also adopt the nuScenes metrics to evaluate the results since the IoU-based metrics are too challenging for camera-based methods. Due to a few camera-based works reported results on Waymo, we also use the official codes of DETR3D to perform experiments on Waymo for comparison. We can observe that BEVFormer outperforms DETR3D by Average Precision with Heading information (APH)~\cite{sun2020scalability} of 6.0\% and 2.5\% on LEVEL\_1 and LEVEL\_2 difficulties with IoU criteria of 0.5. On nuScenes metrics, BEVFormer outperforms DETR3D with a margin of 3.2\% NDS and 5.2\% AP. We also conduct experiments on the front camera to compare BEVFormer with CaDNN~\cite{reading2021categorical}, a monocular 3D detection method that reported their results on the Waymo dataset. BEVFormer outperforms CaDNN with APH of 13.3\% and 11.2\% on LEVEL\_1 and LEVEL\_2 difficulties with IoU criteria of 0.5.

\subsection{Multi-tasks Perception Results}
We train our model with both detection and segmentation heads to verify the learning ability of our model for multiple tasks, and the results are shown in Tab.~\ref{table:multi-tasks}. 
While comparing different BEV encoders under same settings, BEVFormer achieves higher performances of all tasks except for road segmentation results is comparable with BEVFormer-S. For example, with joint training, BEVFormer outperforms Lift-Splat$^*$~\cite{philion2020lift} by 11.0 points on detation task (52.0\% NDS \emph{v.s.} 41.0\% NDS) and IoU of 5.6 points on lane segmentation (23.9\% \emph{v.s.} 18.3\%).
Compared with training tasks individually, multi-task learning  saves computational cost and reduces the inference time by sharing more modules, including the backbone and the BEV encoder. In this paper, we show that the BEV features generated by our BEV encoder can be well adapted to different tasks, and the model training with multi-task heads performs even better on detection tasks and vehicles segmentation. However, the jointly trained model does not perform as well as individually trained models for road and lane segmentation, which is a common phenomenon called \textit{negative transfer}~\cite{crawshaw2020multi,fifty2021efficiently} in multi-task learning.

\begin{table}[t] 
\begin{center}
\caption{\textbf{The detection results of different methods with various BEV encoders on nuScenes \texttt{val} set.} ``Memory'' is the consumed GPU memory during training.
*: We use VPN~\cite{pan2020cross} and Lift-Splat~\cite{philion2020lift} to replace BEV encoder of our model for comparison. ${\text{\textdagger}}$: We train BEVFormer-S using global attention in spatial cross-attention, and the model is trained with fp16 weights. In addition, we only adopt single-scale features from the backbone and set the spatial shape of BEV queries to be $100\!\times\!100$ to save memory.
$\ddag$: We degrade the interaction targets of deformable attention from the local region to the reference points only by removing the predicted offsets and weights.
}

\setlength{\tabcolsep}{1.4mm}
\begin{tabular}{l  c |c c c c |c c c  c }
\toprule
Method &  Attention & NDS$\uparrow$  & mAP$\uparrow$  & mATE$\downarrow$    & mAOE$\downarrow$       & \#Param. &FLOPs& Memory \\
\midrule

VPN$^*$~\cite{pan2020cross}&- & 0.334 &0.252 &0.926 &0.598 & 111.2M &924.5G &$\sim\!$20G\\
List-Splat$^*$~\cite{philion2020lift}&- &  0.397 & 0.348 &0.784 &0.537 &74.0M & 1087.7G&$\sim\!$20G \\

\midrule
\rowcolor{gray95}
BEVFormer-S$^{\text{\textdagger}}$ &Global & 0.404 & 0.325 & 0.837 &0.442&62.1M & 1245.1G& $\sim\!$36G\\ 
\rowcolor{gray9}
BEVFormer-S$^\ddag$& Points &0.423 &0.351 &0.753&0.442  & 68.1M & 1264.3G&$\sim\!$20G\\
\rowcolor{gray85}
BEVFormer-S& Local &\textbf{0.448} &\textbf{0.375}&\textbf{0.725} &\textbf{0.391} & 68.7M &1303.5G &$\sim\!$20G\\
\bottomrule
\end{tabular}
\label{bevformer_s_abl}
\end{center}

\end{table}

\begin{figure}[t]
\centering
\includegraphics[width=\linewidth]{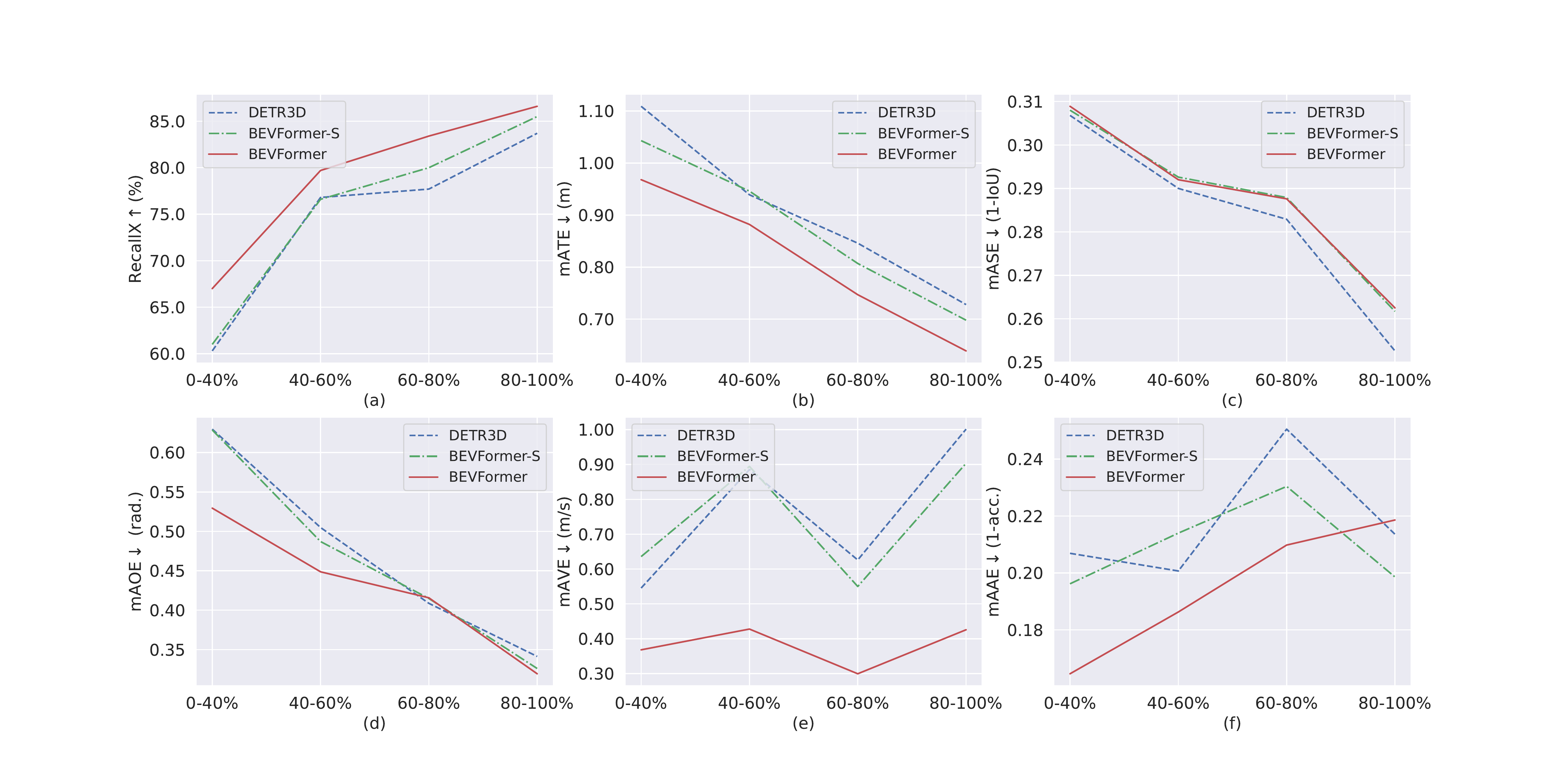}
\caption{\textbf{The detection results of subsets with different visibilities.} We divide the nuScenes \texttt{val} set into four subsets based on the visibility that \{0-40\%, 40-60\%, 60-80\%, 80-100\%\} of objects can be visible. (a): Enhanced by the temporal information, BEVFormer has a higher recall on all subsets, especially on the subset with the lowest visibility (0-40\%). (b), (d) and (e): Temporal information benefits translation, orientation, and velocity accuracy. (c) and (f): The scale and attribute error gaps among different methods are minimal. Temporal information does not work to benefit an object's scale prediction.}
\label{fig:plt}

\end{figure}

\subsection{Ablation Study}

To delve into the effect of different modules, we conduct ablation experiments on nuScenes \texttt{val} set with detection head. More ablation studies are in Appendix.

\noindent\textbf{Effectiveness of Spatial Cross-Attention.}
To verify the effect of spatial cross-attention, we use BEVFormer-S to perform ablation experiments to exclude the interference of temporal information, and the results are shown in Tab.~\ref{bevformer_s_abl}.
The default spatial cross-attention is based on deformable attention.
For comparison, we also construct two other baselines with different attention mechanisms: (1)  Using the global attention to replace deformable attention; (2)  Making each query only interact with its reference points rather than the surrounding local regions, and it is similar to previous  methods~\cite{Roddick2019OrthographicFT,rukhovich2022imvoxelnet}. 
For a broader comparison, we also replace the BEVFormer  with the BEV generation methods proposed by VPN~\cite{pan2020cross} and Lift-Spalt~\cite{philion2020lift}. We can observe that deformable attention significantly outperforms other attention mechanisms under a comparable model scale. 
Global attention consumes too much GPU memory, and point interaction has a limited receptive field.  
Sparse attention achieves better performance because it interacts with a priori determined regions of interest, balancing receptive field and GPU consumption.

\noindent\textbf{Effectiveness of Temporal Self-Attention.} 
From Tab.~\ref{main_det} and Tab.~\ref{table:multi-tasks}, we can observe that BEVFormer outperforms BEVFormer-S with remarkable improvements under the same setting, especially on challenging detection tasks. The effect of temporal information is mainly in the following aspects: (1) The introduction of temporal information greatly benefits the accuracy of the  velocity estimation; (2) The predicted locations and orientations of the objects are more accurate with temporal information; (3) We obtain higher recall on heavily occluded objects since the temporal information contains past objects clues, as showed in Fig.~\ref{fig:plt}. 
To evaluate the performance of BEVFormer on objects with different occlusion levels, we divide the validation set of nuScenes into four subsets according to the official visibility label provided by nuScenes. In each subset, we also compute the average recall of all categories with a center distance threshold of 2 meters during matching. The maximum number of predicted boxes is 300 for all methods to compare recall fairly.
On the subset that only 0-40\% of objects can be visible, the average recall of BEVFormer outperforms BEVFormer-S and DETR3D with a margin of more than 6.0\%.

\noindent\textbf{Model Scale and Latency.} We compare the performance and latency of different configurations in Tab.~\ref{table:latency}. We ablate the scales of BEVFormer in three aspects, including whether to use multi-scale view features, the shape of BEV queries, and the number of layers, to verify the trade-off between performance and inference latency. We can observe that configuration C using one encoder layer in BEVFormer achieves 50.1 \% NDS and reduces the latency of  BEVFormer from the original 130ms to 25ms. Configuration D, with single-scale view features, smaller BEV size, and only 1 encoder layer, consumes only 7ms during inference, although it loses 3.9 points compared to the default configuration. However, due to the multi-view image inputs, the bottleneck that limits the efficiency lies in the backbone, and efficient backbones for autonomous driving deserve in-depth study.
Overall, our architecture can adapt to various model scales and be flexible to trade off performance and efficiency.

\begin{table}[t] 
\begin{center}

\caption{
\textbf{Latency and performance of different model configurations on nuScenes \texttt{val} set.} The latency is measured on a V100 GPU, and the backbone is R101-DCN. The input image shape is $900\!\times\!1600$. ``MS'' notes multi-scale view features.
}
\setlength{\tabcolsep}{1.4mm}
\begin{tabular}{c| c c c | c c  c|c | c c}
\toprule
\multirow{2}{*}{Method}&
\multicolumn{3}{c|}{Scale of BEVFormer}
 &  \multicolumn{3}{c|}{Latency (ms)}&  \multirow{2}{*}{FPS} &\multirow{2}{*}{NDS$\uparrow$} &\multirow{2}{*}{mAP$\uparrow$}\\
&MS&BEV&\#Layer& Backbone & BEVFormer & Head&\\
\midrule
BEVFormer & \cmark&$200\!\times\!200$ & 6 &391 & 130& 19 &1.7&\textbf{0.517}&\textbf{0.416}\\
\midrule
A & \xmark&$200\!\times\!200$ & 6 & 387& 87& 19&1.9&0.511&0.406\\
B & \cmark&$100\!\times\!100$ & 6 & 391& 53& 18&2.0&0.504&0.402\\
C & \cmark&$200\!\times\!200$ & 1 & 391& 25& 19&2.1&0.501&0.396\\
D & \xmark&$100\!\times\!100$ & 1 & 387& \textbf{7}& 18&\textbf{2.3}&0.478&0.374\\
\bottomrule

\end{tabular}
\label{table:latency}
\end{center}
\end{table}

\subsection{Visualization Results}

\begin{figure}[t]
\centering
\includegraphics[width=\linewidth]{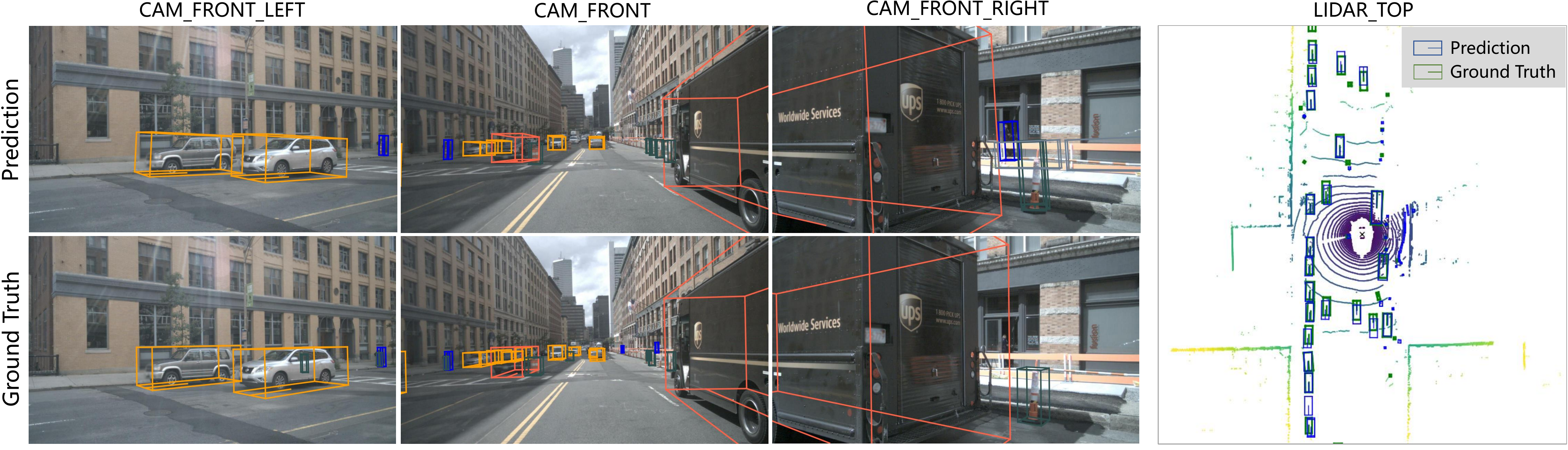}
\caption{\textbf{Visualization results of BEVFormer on nuScenes \texttt{val} set.} We show the 3D bboxes predictions in multi-camera images and the bird's-eye-view. 
}
\label{fig:vis}

\end{figure}

We show the detection results of a complex scene in Fig.~\ref{fig:vis}. BEVFormer produces impressive results except for a few mistakes in small and remote objects. More qualitative results are provided in Appendix.

\section{Discussion and Conclusion}

In this work, we have proposed BEVFormer to generate the bird's-eye-view features from multi-camera inputs. BEVFormer can efficiently aggregate spatial and temporal information and generate powerful BEV features that simultaneously support 3D detection and map segmentation tasks. 

\noindent\textbf{Limitations.} At present, the camera-based methods still have a particular gap with the LiDAR-based methods in effect and efficiency. Accurate inference of 3D location from 2D information remains a long-stand challenge for camera-based methods. 

\noindent\textbf{Broader impacts.}
BEVFormer demonstrates that using spatiotemporal information from the multi-camera input can significantly improve the performance of visual perception models.  The advantages demonstrated by BEVFormer, such as more accurate velocity estimation and higher recall on low-visible objects, are essential for constructing a better and safer autonomous driving system and beyond. We  believe BEVFormer is just a baseline of the following more powerful visual perception methods, and vision-based perception systems still have  tremendous potential to be explored.

\bibliographystyle{splncs04}
\bibliography{egbib}

\clearpage

\section*{Appendix}
\appendix

\section{Implementation Details}
In this section, we provide more implementation details of the proposed method and experiments.

\subsection{Traning Strategy}

Following previous methods~\cite{wang2022detr3d,zhu2020deformable},
we train all models with 24 epochs, a batch size of 1 (containing 6 view images) per GPU, a learning rate of $2\!\times\!10^{-4}$, learning rate multiplier of the backbone is 0.1, and we decay the learning rate with a cosine annealing~\cite{Loshchilov2017SGDRSG}. We employ AdamW~\cite{Loshchilov2019DecoupledWD} with a weight decay of $1\!\times\!10^{-2}$ to optimize our models.

\subsection{VPN and Lift-Splat}
We use VPN~\cite{pan2020cross} and Lift-Splat~\cite{philion2020lift} as two baselines in this work. The backbone and the task heads are the same as the BEVFomer for fair comparisons.

\noindent\textbf{VPN.} We employ the official codes\footnote{https://github.com/pbw-Berwin/View-Parsing-Network} in this work. Limited by the huge amount of parameters of MLP, it is difficult for VPN to generate high-resolution BEV (\eg, 200$\times$ 200).
To compare with VPN, in this work, we transform the single-scale view features into BEV with a low resolution of $50\!\times\!50$ 
via two view translation layers.

\noindent\textbf{Lift-Splat.} We enhance the camera encoder of Lift-Splat\footnote{https://github.com/nv-tlabs/lift-splat-shoot} with two additional convolutional layers for a fair comparison with our BEVFormer under a comparable parameter number. Other settings remain unchanged.

\subsection{Task Heads} 



\begin{wrapfigure}{r}{0.35\linewidth}
    \vspace{-35px}
    \renewcommand{\captionlabelfont}{\scriptsize}
    \begin{center}
    \includegraphics[width=\linewidth]{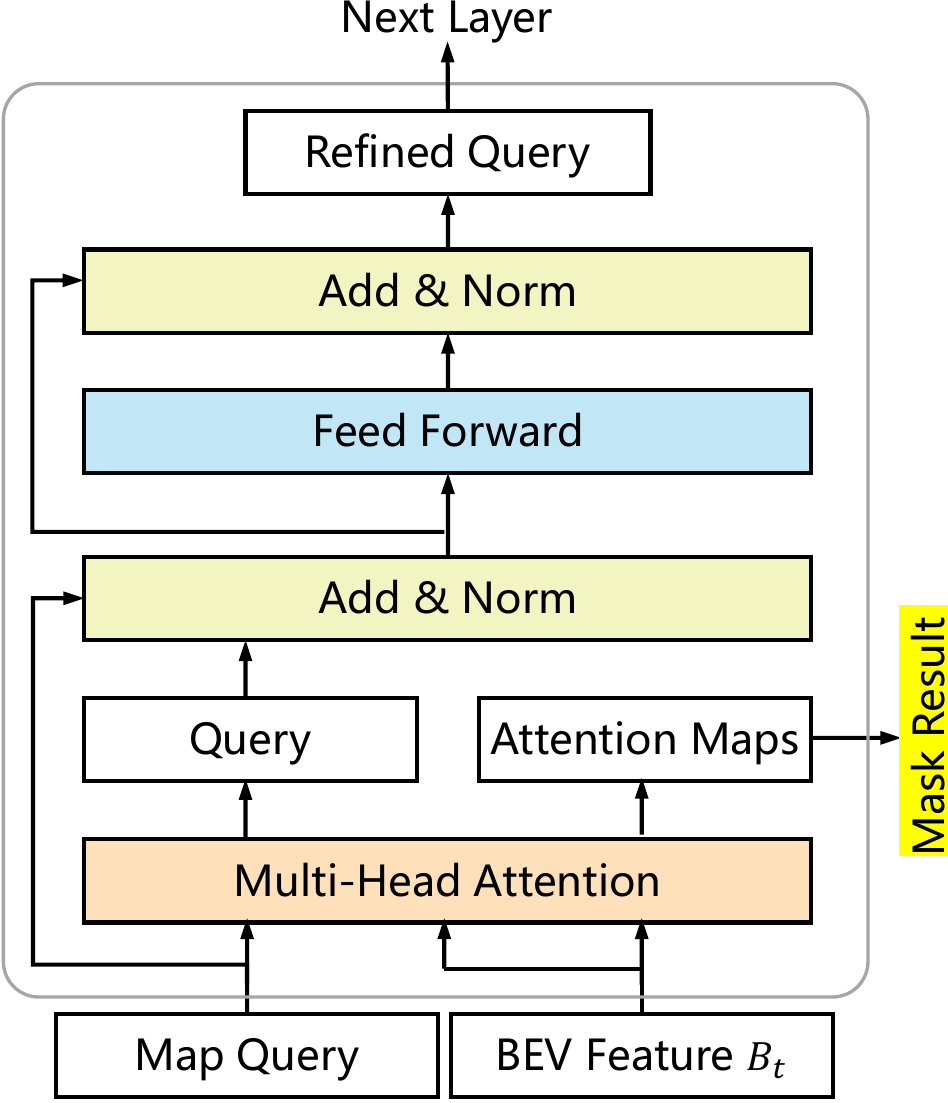}
    \caption{\scriptsize \textbf{
    Segmentation head (mask decoder) of BEVFormer.
    }}
    \label{fig:appdendix_arch}
    \end{center}
    \vspace{-40px}
\end{wrapfigure}
\textbf{Detection Head.} We predict 
10 parameters for each 3D bounding box, 
including the 3 parameters  $(l,w,h)$ for the scale of each box, 3 parameters $(x_o,y_o,z_o)$ for the center location, 2 parameters  (cos($\theta$), sin($\theta$)) for object's yaw $\theta$, 2 parameters $(v_x,v_y)$ for the  velocity. Only $L_1$ loss and $L_1$ cost are used during training phase. Following ~\cite{wang2022detr3d}, we use 900 object queries and keep 300 predicted boxes with highest confidence scores during inference.

\noindent\textbf{Sementation Head.} 
As shown in Fig.~\ref{fig:appdendix_arch}, for each class of the semantic map, we follow the mask decoder in \cite{li2021panoptic} to use one learnable query to represent this class, and generate the final segmentation masks based on the attention maps from the vanilla multi-head attention.

\subsection{Spatial Cross-Attention}

\noindent\textbf{Global Attention.} Besides deformable attention~\cite{zhu2020deformable}, our spatial cross-attention can also be implemented by global attention (\ie,  vanilla multi-head attention)~\cite{vaswani2017attention}. The most straightforward way to employ global attention is making each BEV query interact with all multi-camera features, and this conceptual implementation does not require camera calibration.
However, the computational cost of this straightforward way is unaffordable. Therefore, we still utilize the camera intrinsic and extrinsic to decide the hit views that one BEV query deserves to interact. This strategy makes that one BEV query usually interacts with only one or two views rather than all views, making it possible to use global attention in the spatial cross-attention. Notably, compared to other attention mechanisms that rely on precise camera intrinsic and extrinsic, global attention is more robust to camera calibration.

\section{Robustness on Camera Extrinsics}
\begin{figure}[t]
\centering

\includegraphics[width=0.75\linewidth]{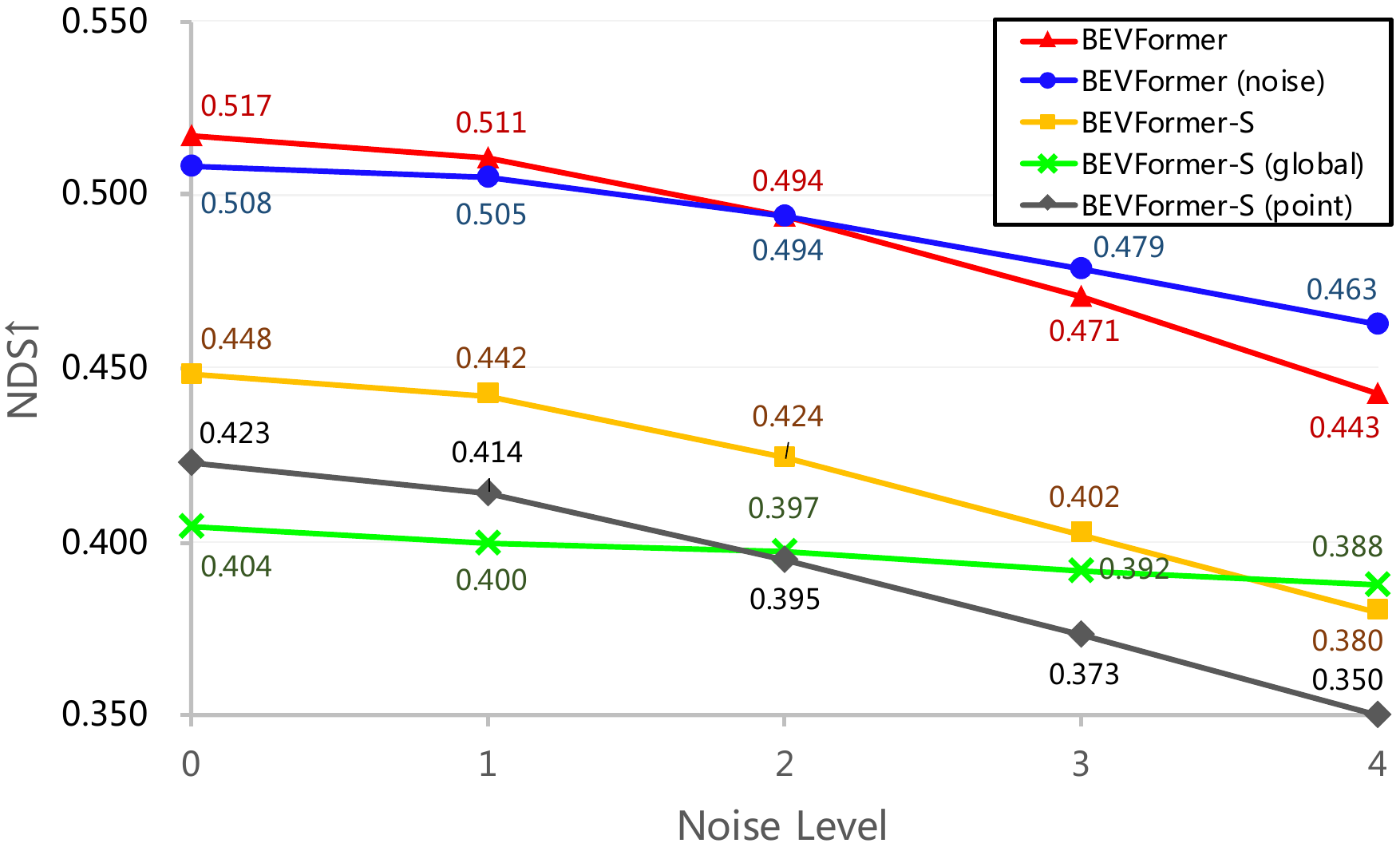}
\caption{\textbf{NDS of methods on nuScenes val set subjected to different levels of camera extrinsics noises.} For $i$-th level noises, the rotation noises are sampled from a normal distribution with mean equals 0 and variance equals $i$ (rotation noise are in degrees, and the noise of each axis is independent), and the translation noises are sampled from a normal distribution with mean equals 0 and variance equals $5i$ (translation noises are in centimeters,  and the noise of each direction is independent).
``BEVFormer'' is our default version.
``BEVFormer (noise)'' is trained with noisy extrinsics (noise level=1). 
``BEVFormer-S'' is our static version of BEVFormer with the spatial cross-attention implemented by deformable attention~\cite{zhu2020deformable}. ``BEVFormer-S (global)'' is BEVFormer-S with the spatial cross-attention implemented by global attention (\ie, vanilla multi-head attention)~\cite{vaswani2017attention}. ``BEVFormer-S (point)'' is BEVFormer-S with point spatial cross-attention where we degrade the interaction targets of deformable attention from the local region to the reference points only by removing the predicted offsets and weights. }
\label{fig:noise}

\end{figure}

BEVFormer relies on camera intrinsics and extrinsics to obtain the reference points on 2D views. During the deployment phase of autonomous driving systems, extrinsics may be biased due to various reasons such as calibration errors, camera offsets, etc. As shown in Fig.~\ref{fig:noise}, we show the results of models under different camera extrinsics noise levels.
Compared to BEVFormer-S (point), BEVFormer-S utilizes the spatial cross-attention based on deformable attention~\cite{zhu2020deformable} and samples features around the reference points rather than only interacting with the reference points. 
With deformable attention, the robustness of 
BEVFormer-S  is stronger than BEVFormer-S (point). For example, with the noise level being 4, the NDS of BEVFormer-S drops 15.2\% (calculated by $1-\frac{0.380}{0.448}$), while the NDS of BEVFormer-S (point) drops 17.3\%. Compared to BEVFormer-S, BEVFormer only drops 14.3\% NDS, which shows that temporal information can also improve robustness on camera extrinsics.
Following \cite{philion2020lift}, we show that when training BEVFormer with noisy extrinsics, BEVFormer (noise) has stronger robustness (only drops 8.9\% NDS). With the spatial cross-attention based on global attention, BEVFormer (global) has a strong anti-interference ability (4.0\% NDS drop) even under level 4 of the camera extrinsics noise. The reason is that we do not utilize camera extrinsics to select the RoIs for BEV queries.

Notably, under the harshest noises, we see that BEVFormer-S (global) even outperforms BEVFormer-S (38.8\% NDS \emph{vs.} 38.0\% NDS). 

\section{Ablation Studies}

\noindent\textbf{Effect of the frame number during training.} Tab.~\ref{tab:table1_a} shows the effect of the frame number during training. We see that the NDS on nuScenes val set keeps
rising with the growth of the frame number and begins to level off the frame number $\ge 4$. Therefore, we set the frame number during training to 4 by default in experiments.

\noindent\textbf{Effect of some designs.} Tab.~\ref{tab:table1_b} shows the results of several ablation studies. Comparing \#1 and \#4, we see that aligning history BEV features with ego-motion is important to represent the same geometry scene as current BEV queries (51.0\% NDS \emph{vs.} 51.7\% NDS). Comparing \#2 and \#4, randomly sampling 4 frames from 5 frames is a effective data augment strategy to improve performance (51.3\% NDS \emph{vs.} 51.7\% NDS). Compared to only using the BEV query to predict offsets and weights during the temporal self-attention module (see \#3), using both BEV queries and history BEV features (see \#4) contain more clues about the past BEV features and benefits location prediction (51.3\% NDS \emph{vs.} 51.7\% NDS).

\begin{table}

\begin{minipage}[!t]{0.5\linewidth}

\centering
\captionsetup{width=.95\textwidth}
\caption{\textbf{NDS of models on nuScenes val set using different frame numbers during training.} ``\#Frame'' denotes the frame number during training.
}
\begin{center}

\setlength{\tabcolsep}{2.0mm}
\begin{tabular}[t]{c| c c c}
\toprule

\#Frame & NDS$\uparrow$ & mAP$\uparrow$ &mAVE$\downarrow$\\
\midrule
1 & 0.448 & 0.375 &0.802\\ 
2 & 0.490 &0.388 & 0.467\\
3 & 0.510 & 0.410 &0.423 \\
4 & \textbf{0.517} & \textbf{0.416} & 0.394\\
5 & \textbf{0.517} & 0.412 & \textbf{0.387}\\
\bottomrule
\end{tabular}
\label{tab:table1_a}
\end{center}
\end{minipage}%
\begin{minipage}[!t]{0.5\linewidth}
\centering
\captionsetup{width=.95\textwidth}
\caption{\textbf{Ablation Experiments on nuScenes val set.} ``A.'' indicates aligning history BEV features with ego-motion. ``R.'' indicates randomly sampling 4 frames from 5 continuous frames. ``B.'' indicates using both BEV queries and history BEV features to predict offsets and weights.}
\renewcommand{\arraystretch}{0.88}
\setlength{\tabcolsep}{0.5mm}
\begin{center}
\addtolength{\tabcolsep}{4pt}
\begin{tabular}[t]{c | c c c |c c}
\toprule

\#   & A. & R. & B. & NDS$\uparrow$ & mAP$\uparrow$ \\
\midrule
1 & \xmark & \cmark & \cmark & 0.510& 0.410\\
2  & \cmark & \xmark & \cmark &0.513 & 0.410\\
3  & \cmark & \cmark & \xmark &  0.513 & 0.404\\
\midrule
4 &\cmark & \cmark & \cmark &  \textbf{0.517} & \textbf{0.416}\\
\bottomrule
\end{tabular}

\label{tab:table1_b}
\end{center}
\end{minipage}

\label{tab:table1}

\end{table}

\section{Visualization}

\begin{figure}[h]
\centering
\includegraphics[width=\linewidth]{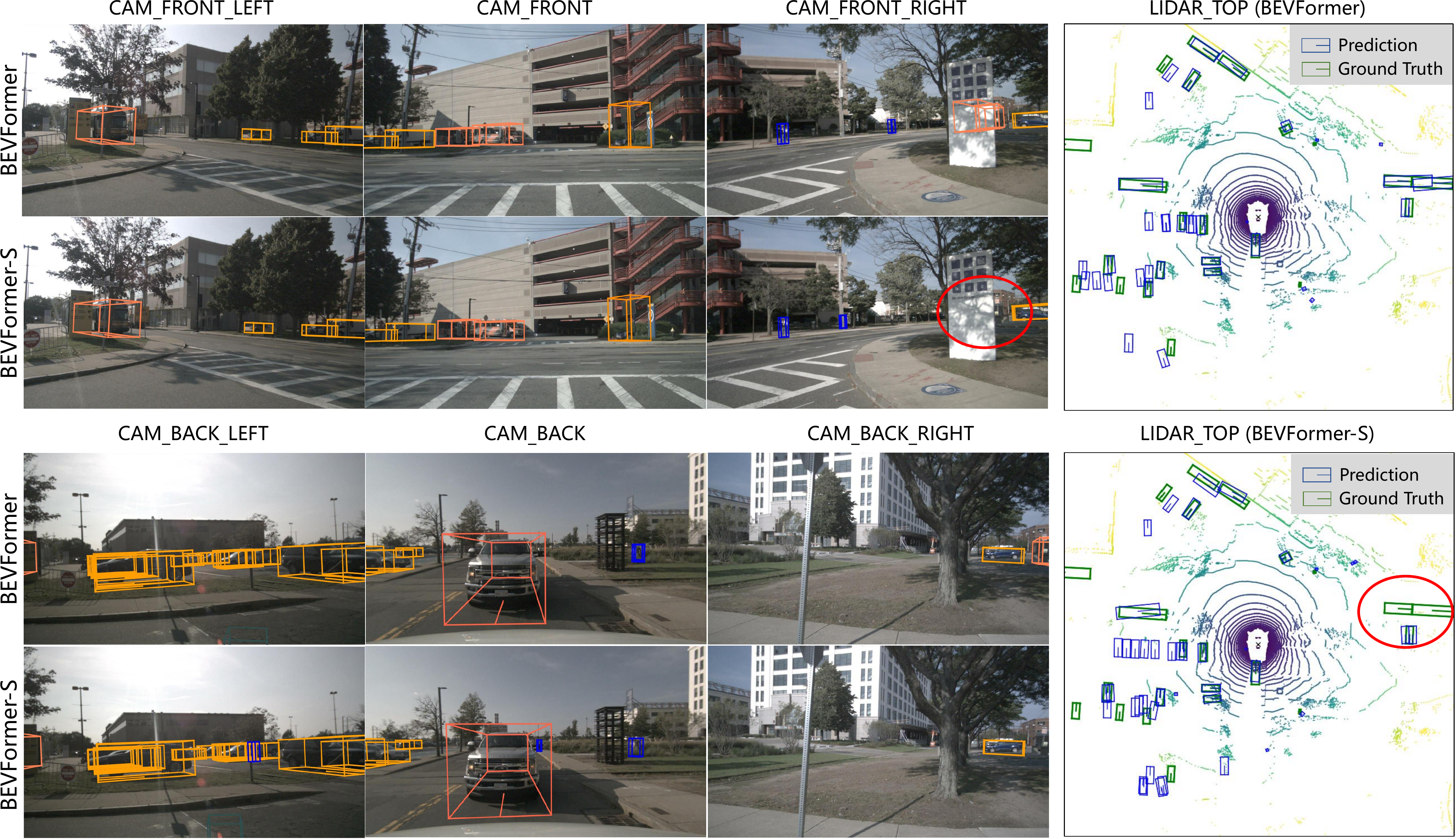}
\caption{\textbf{Comparision of BEVFormer and BEVFormer-S on nuScenes val set.} We can observe that BEVFormer can detect highly occluded objects, and these objects are missed in the  prediction results of BEVFormer-S (in red circle). 
}
\label{fig:static_vs_video}
\end{figure}

\begin{figure}[b]
\centering
\includegraphics[width=\linewidth]{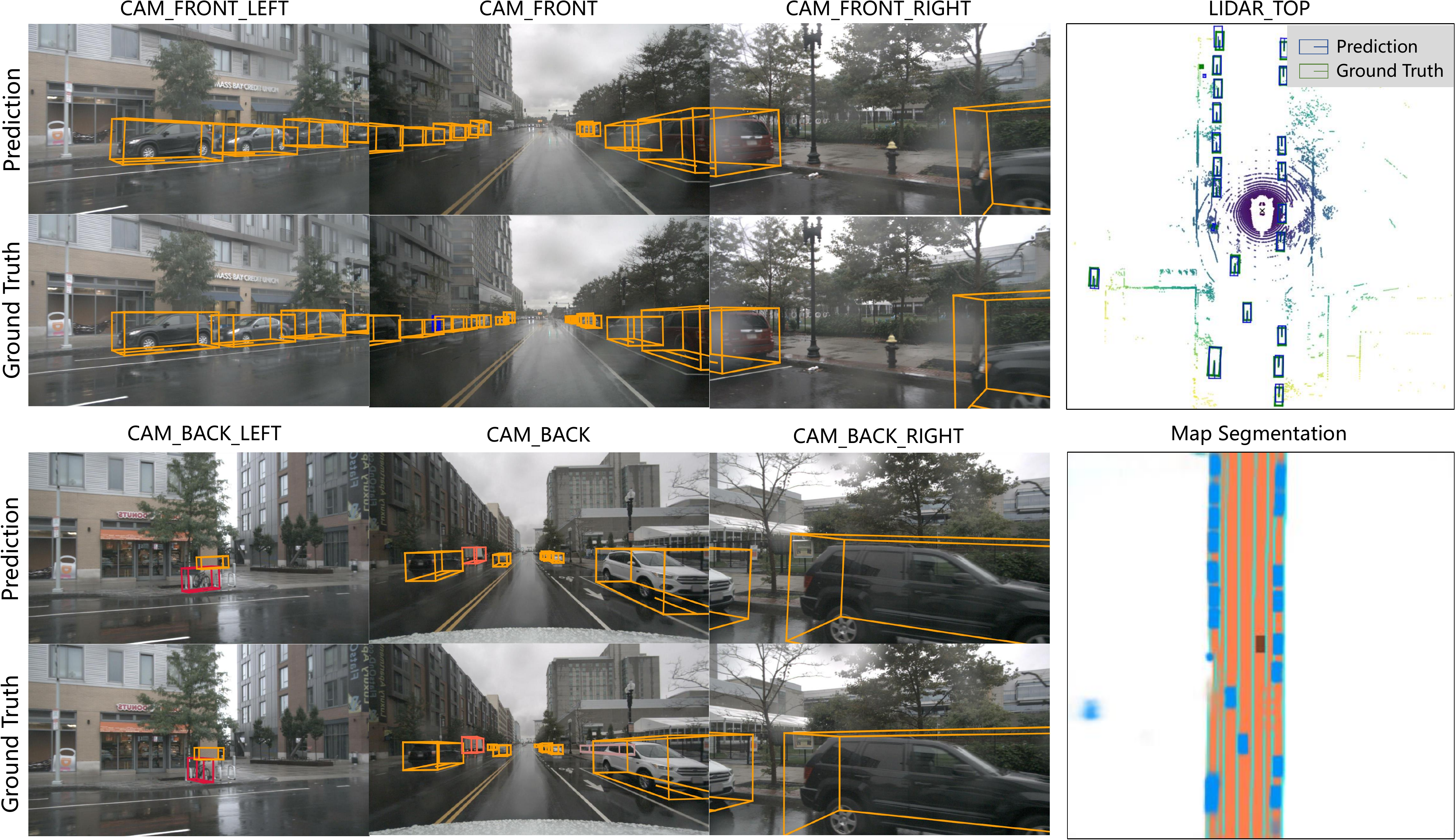}
\caption{\textbf{Visualization results of both object detection and map segmentation tasks.} We show vehicle, road, and lane segmentation in blue, orange, and green, respectively. 
}
\label{fig:visual_4000}
\end{figure}

\begin{figure}[h]
\centering
\includegraphics[width=\linewidth]{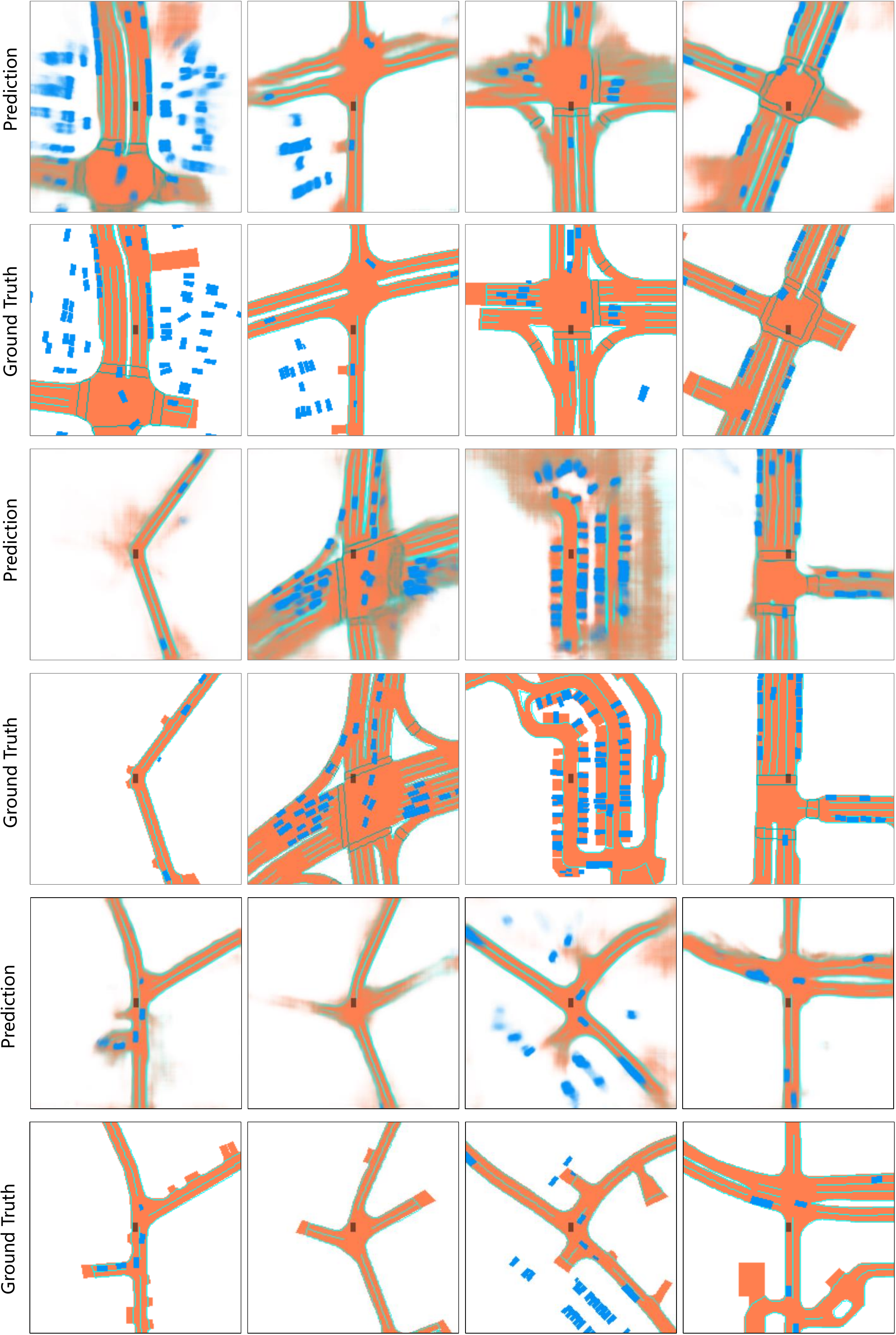}
\caption{\textbf{Visualization results of the map segmentation task.} We show vehicle, road, ped crossing and lane segmentation in blue, orange,  cyan, and green, respectively. 
}
\label{fig:mask}
\end{figure}

As shown in Fig.~\ref{fig:static_vs_video},  we compare BEVFormer with BEVFormer-S. With temporal information, BEVFormer successfully detected two buses occluded by  boards. We also show both object detection and map segmentation results in Fig.~\ref{fig:visual_4000}, where we see that the detection results and segmentation results are highly consistent. We provide more map segmentation results in Fig.~\ref{fig:mask}, where we see that with the strong BEV features generated by BEVFormer, the semantic maps can be well predicted via a simple mask decoder.
\vspace{16mm}

\end{document}